\documentclass[letterpaper, 10 pt, conference]{ieeeconf}  

\IEEEoverridecommandlockouts                              

\usepackage{amsmath}
\usepackage{amssymb}

\usepackage{hyperref}
\hypersetup{
    colorlinks=true,
    linkcolor=blue,
    filecolor=magenta,      
    urlcolor=cyan,
    breaklinks=true,
}

\makeatletter
\g@addto@macro{\UrlBreaks}{\UrlOrds}
\makeatother

\usepackage{subfig}
\usepackage{gensymb}
\newsavebox{\bigleftbox}

\usepackage{makecell}

\usepackage{tikz}  
\usetikzlibrary{spy}
\usepackage{color, colortbl}
\usepackage{pgfplots}
\pgfplotsset{compat=newest}
\usepackage{multicol}
\usepackage{circuitikz}
\usepackage[nolist]{acronym}
\usepgfplotslibrary{groupplots}
\usetikzlibrary{shapes}
\usepackage{siunitx}  
\usepackage{graphicx}

\definecolor{mittelblau}{RGB}{0, 126, 198}
\definecolor{violettblau}{cmyk}{0.9, 0.6, 0, 0}
\definecolor{rot}{RGB}{238, 28 35}
\definecolor{apfelgruen}{RGB}{140, 198, 62}
\definecolor{gelb}{RGB}{1, 221, 0}
\definecolor{orange}{RGB}{244, 111, 33}
\definecolor{pink}{RGB}{237, 0, 140}
\definecolor{lila}{RGB}{128, 10, 145}
\definecolor{hellgrau}{RGB}{224, 224, 224}
\definecolor{mittelgrau}{RGB}{128, 128, 128}
\definecolor{dunkelgrau}{RGB}{80,80,80}
\definecolor{anthrazit}{RGB}{19, 31, 31}
\definecolor{darkgreen}{RGB}{0.125,0.5,0.169}

\usepackage{ifthen}
\usetikzlibrary{matrix,calc}

\usepackage[colorinlistoftodos]{todonotes}

\usepackage[ruled]{algorithm2e}

\title{\LARGE \bf
Deep Inertial Navigation using \\Continuous Domain Adaptation and Optimal Transport
}

\author{
    \authorblockN{
        Mohammed Alloulah\authorrefmark{1}\authorrefmark{2}, 
        Maximilian Arnold\authorrefmark{2}, 
        Anton Isopoussu\authorrefmark{3},
    }
    \authorblockA{
        \authorrefmark{2}Bell Labs
        \hspace{0.25cm} 
        \authorrefmark{3}Invenia Labs
    }
}

\newcommand\blfootnote[1]{%
  \begingroup
  \renewcommand\thefootnote{}\footnote{#1}%
  \addtocounter{footnote}{-1}%
  \endgroup
}

\begin{document}

\maketitle
\thispagestyle{empty}
\pagestyle{empty}

\begin{abstract}
In this paper, we propose a new strategy for learning inertial robotic navigation models.
The proposed strategy enhances the generalisability of end-to-end inertial modelling, and is aimed at wheeled robotic deployments.
Concretely, the paper describes the following.
(1) Using precision robotics, we empirically characterise the effect of changing the sensor position during navigation on the distribution of raw inertial signals, as well as the corresponding impact on learnt latent spaces.
(2) We propose neural architectures and algorithms to assimilate knowledge from an \emph{indexed} set of sensor positions in order to enhance the robustness and generalisability of robotic inertial tracking in the field.
Our scheme of choice uses continuous domain adaptation (DA) and optimal transport (OT).
(3) In our evaluation, continuous OT DA outperforms a continuous adversarial DA baseline, while also showing quantifiable learning benefits over simple data augmentation.
We will release our dataset to help foster future research.
\end{abstract}


\begin{multicols}{2}
\begin{acronym}[WSSUS]
 \acro{5G}{fifth-generation}
 \acro{ADC}{analog to digital converter}
 \acro{AFE}{analog front end}
 \acro{AGC}{automatic gain control}
 \acro{AGV}{automated guided vehicle}
 \acro{AMP}{approximate message passing}
 \acro{AWGN}{additive white Gaussian noise}

 \acro{BER}{bit error rate}

 \acro{BB}{baseband}
 \acro{bpcu}{bits per channel use}
 \acro{BP}{belief propagation}
 \acro{BPSK}{binary phase shift keying}
 \acro{BS}{base station}

 \acro{CB}{codebook}
 \acro{CDF}{cumulative distribution function}
 \acro{CFO}{carrier frequency offset}
 \acro{CoSaMP}{compressive sampling matching pursuit}
 \acro{CP}{cyclic prefix}
 \acro{CS}{compressive sensing} 
 \acro{CSI}{channel state information}
 \acro{CNN}{convolutional neural network}

\acro{DA}{domain adaptation}
 \acro{DAC}{digital-analog-converter}
 \acro{DC}{direct current}
 \acro{DE}{distance error}
 \acro{DeepL}{deep-learning}
 \acro{DoF}{degree-of-freedom}
 \acro{DFT}{discrete Fourier transformation}
 \acro{DL}{deep learning}
 \acro{DS}{delay spread}
 \acro{DSP}{digital signal processing}

 \acro{ECC}{error-correcting code}
 \acro{ENoB}{effective number of bits}
 \acro{ERP}{effective radiated power}
 \acro{EVM}{error vector magnitude}

 \acro{FB}{feedback}
 \acro{FC}{fully connected}
 \acro{FDD}{frequency division duplexing}
 \acro{FDM}{frequency division multiplexing}
 \acro{FIR}{finite impulse response}
 \acro{FT}{fine tuning}
 \acro{FPGA}{field programmable gate array}

\acro{GAN}{Generative adversarial network}
 \acro{GPIO}{general-purpose input/output}
 \acro{GPS}{global positioning system}
 \acro{GPSDO}{GPS disciplined oscillator}
 \acro{GPU}{graphical processing unit}

 \acro{HDD}{hard decision decoding}

 \acro{IC}{integrated circuit}
 \acro{I2C}{Inter-Integrated Circuit}
 \acro{ICSP}{in-circuit serial programming}
 \acro{IF}{intermediate frequency}
 \acro{i.i.d.}{independent and identically distributed}
 \acro{IIR}{infinite impulse response}
 \acro{IMU}{inertial measurement unit}
 \acro{IoT}{Internet of Things}
 \acro{IPS}{indoor positioning system}
 \acro{IR}{infrared}
 \acro{JSDM}{Joint Spatial Division and Multiplexing}

 \acro{LLR}{log-likelihood ratio}
 \acro{LP}{leakage precoder}
 \acro{LMMSE}{Linear Minimum Mean Square Error}
 \acro{LO}{local oscillator}
 \acro{LoS}{line of sight}

 \acro{LiDaR}{Light Detection and Ranging}
 \acro{LS}{least squares}
 \acro{LSTM}{long-term short-term memory}
 \acro{LTE}{Long Term Evolution}
 \acro{LTI}{linear time invariant}
 \acro{LTV}{linear time variant}
  
 \acro{MAP}{maximum a posteriori}
 \acro{MDE}{mean distance error}
 \acro{MDA}{mean distance accuracy}
  \acro{MEMS}{Micro-Electro-Mechanical Systems}
 \acro{MIMO}{multiple input multiple output}
 \acro{MISO}{multiple input single output}
 \acro{ML}{maximum likelihood}
 \acro{MLD}{maximum likelihood decoding}
 \acro{mMIMO}{massive multiple input multiple output}
 \acro{MMSE}{minimum mean square error}
 \acro{M-MMSE}{multi-cell minimum mean square error}
 \acro{MR}{maximum ratio}
 \acro{MRC}{maximum ratio combining}
 \acro{MRP}{maximum ratio precoding}
 \acro{MRT}{maximum ratio transmission}
 \acro{MSE}{mean squared error}
 \acro{MQTT}{Message Queuing Telemetry Transport}
 \acro{MU}{multi-user}

 \acro{NF}{noise figure}
 \acro{NN}{Neural Network}
 \acro{NNI}{Neural Network Intelligence}
 \acro{NLoS}{non-line of sight}
 \acro{NND}{neural network decoding}
  \acro{NTP}{Network Time Protocol}
 \acro{NMSE}{normalized mean squared error}
 \acro{NU}{not-used}

 \acro{OFDM}{orthogonal frequency division multiplex}
 \acro{OMP}{orthogonal matching pursuit}
 \acro{OPS}{outdoor positioning system}
 \acro{OT}{optimal transport}
 
 \acro{PB}{passband}
 \acro{PCB}{printed circuit board}
 \acro{PDR}{pedestrian dead reckoning}
 \acro{PDF}{probability density function}
 \acro{PDP}{power-delay-profile}
\acro{PLL}{phase-locked-loop}
 \acro{PO}{phase-only}
 \acro{PPS}{pulse per second}

 \acro{QPSK}{quadrature phase shift keying}
 \acro{QuaDRIGa}{Quasi Deterministic Radio Channel Generator}

 \acro{ReLU}{rectified linear unit}
 \acro{RF}{radio frequency}
 \acro{RMS-DS}{Root Mean Square - Delay Spread}
 \acro{RNN}{recurrent neuronal network}
 \acro{RSSI}{received signal strength indicator}
 \acro{R-ZF}{regularized zero-forcing}

 \acro{SDD}{soft decision decoding}
 \acro{SDR}{software defined radio}
 \acro{SE}{spectral efficiency}
 \acro{SFO}{sampling frequency offset}
 \acro{SLAM}{Simultaneous Localization and Mapping}
 \acro{SGD}{stochastic gradient descent}
 \acro{SISO}{single input single output}
 \acro{SINR}{signal-to-interference-and-noise-ratio}
 \acro{SIR}{signal-to-interference-ratio}
 \acro{SLNR}{signal-to-leakage-and-noise ratio}
 \acro{SNR}{signal-to-noise-ratio}
 \acro{SP}{subspace}
 \acro{SQR}{signal-to-quantization-noise-ratio}
 \acro{SQNR}{signal-to-quantization-noise-ratio}
 \acro{SVD}{singular value decomposition}
 \acro{SU}{single-user}

 \acro{TDD}{time division duplexing}
 \acro{TRIPS}{time-reversal IPS}

 \acro{UE}{user equipment}
 \acro{UL}{uplink}
 \acro{ULA}{uniform line array}
 \acro{URLLC}{ultra-reliable low-latency communication}
 \acro{US}{uncorrelated scattering}
 \acro{USRP}{universal software radio peripheral}
  \acro{WiFi}{Wireless Fidelity}
 \acro{WSS}{wide sense stationary}
 \acro{WSSUS}{wide sense stationary uncorrelated scattering}

 \acro{ZF}{zero forcing}
\end{acronym}
\end{multicols}

\vspace{-1cm}
\section{Introduction}\label{Sec:Introduction}

A\blfootnote{\authorrefmark{1}Correspondence to \href{mailto:alloulah@outlook.com}{alloulah@outlook.com}} wide array of sensing modalities has enabled important and iconic location-based services, both indoors and outdoors. 
More recent research aspires to support autonomous navigation in uninstrumented and unprepared indoor environments \cite{Ayyala20_DLoc}.
Such infrastructure-less navigation could bring about a new class of location-based services, for example the fully autonomous factory~\cite{Brossard20_AI-IMU}.
Extrapolating from the outdoors, this parallel indoor ambition will in no doubt also require collaborative sensing modalities in order to sustain performance irrespective of challenging dynamic conditions.
Of the many modalities on offer, inertial tracking plays a key role under momentary unfavorable conditions such as occlusion or low \ac{SNR}. 
Because of its independence of the surrounding environment, inertial navigation can help maintain location estimates and significantly shrink the error long tail of single-modal systems~\cite{Liu20_Tlio}.

Inertial tracking has traditionally suffered from excessive error growth and required extensive and cumbersome tuning~\cite{Herath20_Ronin,Brossard20_AI-IMU,Liu20_Tlio}. 
Recent advances recast inertial tracking as a sequence-based learning task~\cite{Chen19_DnnInertialOdometry}. 
Data-driven inertial tracking models, such as \ac{DL}, can demonstrably capture and compensate for error growths.
However, inertial signals remain overly prone\footnote{when compared to other sensing modalities such as vision and sound} to changes in the data distribution that are hard to control for when deployed in the field.
There are a number of factors that give rise to these distributional shifts.
For example, the motion profiles IMUs experience on a wheeled robot are fundamentally different when compared to a pedestrian. 
Changes in the position of the sensor on a moving object also cause significant distributional shifts~\cite{Chen19_DnnInertialOdometry,Liu20_Tlio}.
Disregarding the underlying causes, prior art treats inertial data shift as a categorical domain shift problem~\cite{Chen19_MotionTransformer}.
Categorical domain adaptation (DA) maps a model trained on a single signal distribution to another---more conveniently without labels---in order to enhance the model's potential for generalisation under \emph{uncontrolled} deployments~\cite{Wilson20_DaSurvey}. 
A more recent DA approach seeks to perform adaptation across a continuously-indexed set of domains in order to capture the underlying collective structure of data, such as disease proneness as a function of age~\cite{Wang20_CIDA}. 
That is, it is not unreasonable to postulate that the total dynamics of age can inform the task of disease diagnosis for any specific age ``domain''.

In this work we focus mainly on inertial navigation for wheeled robots.
We want to support industrial automation applications, such as food delivery by autonomous robots during the COVID-19 lockdown~\cite{Guardian_RobotsDeliverFood}.
This use case has seen less attention in recent breakthrough data-driven inertial navigation research~\cite{Herath20_Ronin,Liu20_Tlio,Chen19_DnnInertialOdometry}, which has mainly dealt with \emph{pedestrian motions}.
Our overarching goal is to \emph{automatically} match a robot in the field (i.e. without any prior knowledge of its construction) to a pre-trained navigation model, thereby enhancing the support for localisation as a service for 3rd party robots.
Taking first steps towards this ambition, we aim to understand systematically the underpinnings of inertial signal variabilities due to changes in sensor position in \emph{wheeled robotic motions}.
We thus ask two fundamental research questions:
\begin{itemize}
  \item What is the effect of different sensor positions on inertial signals?
  \item Can sensor position \emph{diversity}\footnote{i.e. inertial signals sampled from many inter-related sensor positions} help inertial DL generalise better?
\end{itemize}

\begin{figure*}[t]
\centering
  \subfloat[]{
    \hspace{-0.425cm}
    \includegraphics{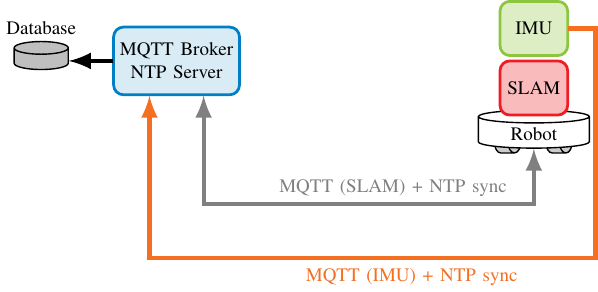}
    \label{fig:Schematic}
  }   
  \subfloat[]{
    \includegraphics{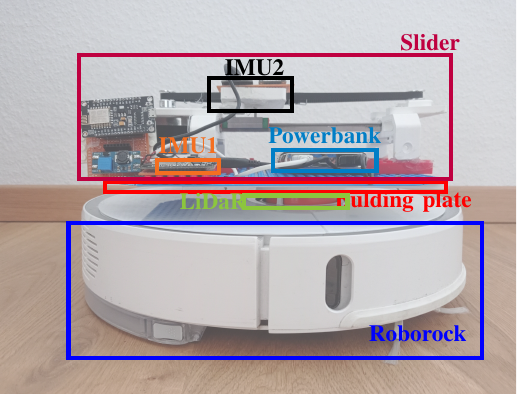}
    \label{fig:SidePictureRoborock}
  }
  \subfloat[]{
    \hspace{-0.385cm}
    \includegraphics{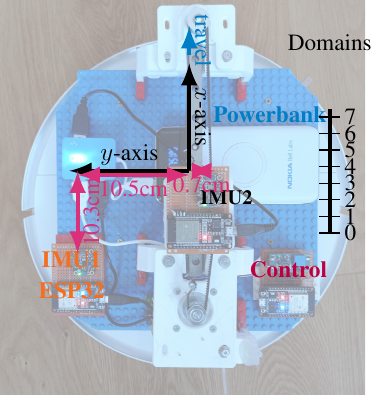}
    \label{fig:TopPictureRoborock}
  }
  \vspace{-0.15cm}
  \caption{\protect\subref{fig:Schematic} Block diagram of data collection system.
  \protect\subref{fig:SidePictureRoborock} Side view of in-house fabricated platform for empirical measurement campaign.
  \protect\subref{fig:TopPictureRoborock} Annotated top view of the platform showing the geometric relationship of indexed IMU domains.
  } 
  \label{fig:measurement_system}
  \vspace{-0.4cm}
\end{figure*}

In this paper, we set out to answer the above research questions. 
We propose to learn a robust navigation model using data enriched with inertial characteristics from many sensor positions. 
Specifically, our approach is to a) mechanically increment the sensor position on a robot, then b) capture and learn this ``robotically generated'' inertial domain in a data-driven fashion.
Our intuition is to treat these sensor position-induced inertial variabilities as an asset for learning and not a nuisance parameter.
Concretely, our contributions are threefold.

\noindent \textbf{(i)} We build the \emph{xSlider}: A robotic platform equipped with a mechanical slider on which an IMU is mounted. 
The mechanical slider induces precise positional shifts to the mounting location of the IMU on the robot. 
The data xSlider generates is therefore \emph{indexed} according to the position of the sensor. 
The robotic platform also supplies simultaneous localisation and mapping (SLAM) measurements that are used as labels for training.
Using the robotic platform, we curate in-house the \emph{xSlider} dataset: a 7.2 million sample IMU data covering an aggregate distance of 20 kilometres split into 8 indexed sensor positions.

\noindent \textbf{(ii)} We design a neural architecture for displacement and heading estimation. The neural architecture combines local convolutional summaries with long recurrent context tracking.
The architecture incorporates algorithms that are able to assimilate knowledge from the robotically-induced inertial domains xSlider supplies it with.

\noindent \textbf{(iii)} We extensively characterise our proposed neural architecture and algorithms against a number of state-of-the-art baselines.

We will release our indexed dataset and CAD designs for xSlider in order to foster further research towards the pursuit of full robotic autonomy.

\vspace{-0.2cm}
\section{xSlider Dataset}

The xSlider dataset is designed to study the implications of sensor position on deep inertial navigation for wheeled robots. Wheeled robotics are an important enabler for next generation industrial automation applications.

The dataset features a series of precision sensor position experiments. We mount an \ac{IMU} on a wheeled robot equipped with a groundtruth SLAM localisation system.\footnote{Details of our experimental testbed will be given in a supplementary material.}
Specifically, the \ac{IMU} position w.r.t. the robot is made programmable using a mechanical slider subsystem (see Fig.~\ref{fig:SidePictureRoborock}).
As such, the centrifugal forces the \ac{IMU} experiences during robot navigation (i.e. while turning) will vary as a function of slider run-time configurations.
This is conceptually similar to how a centrifuge simulator for building high-G tolerance in astronauts works~\cite{Wikipedia_High-GTraining}.
We operate the robot in an indoor space and collect \ac{IMU} measurements accompanied with pseudo groundtruth coordinate labels.
We repeat the experiment for 8 sessions, with each session lasting approx. 6 hours.
For each session, we increment the position of the \ac{IMU} on the slider by $1$cm.
The combined slider positions amount to exactly \SI{7}{\centi\metre} translation of mechanically-indexed IMU domains.

\begin{table}[t]
\centering
\footnotesize
\begin{tabular}{ |p{2.2cm}||p{2.4cm}|p{2.7cm}|  }
 \hline
 \multicolumn{1}{|c||}{\textbf{Dataset}} &\multicolumn{1}{c|}{\textbf{xSlider}} &\multicolumn{1}{|c|}{\textbf{OxIOD}} \\
 \hline
 Source             &In-house       &3rd party              \\
 Year               &2021           &2018                   \\
 Sampling (Hz)      &70             &100                    \\
 Labels             &SLAM           &Vicon                  \\
 Labels dimensions  &2D             &3D                     \\
 Labels accuracy    &$<$1cm         &$<$1mm                 \\
 Domains            &8 indexed      &4+2=6 categorical      \\
 Domain control     &automated      &manual                 \\
 Paired domain      &yes            &no                     \\
 Type               &wheeled robot  &human attachments      \\
 IMU                &Bosch BNO055   &InvenSense ICM-20600   \\
 Size (km)          &2$\times$ 19.512 &42.587               \\
 Time (hour)        &54             &14.7                   \\
 \hline
\end{tabular}
\caption{Indoor multi-domain IMU datasets for \ac{DL}.}
\label{table:dl_imu_datasets}
\vspace{-0.4cm}
\end{table}

Tab.~\ref{table:dl_imu_datasets} compares xSlider to OxIOD~\cite{Chen18_OxIOD} which is a closest \emph{multi-domain} alternative dataset.

\vspace{-0.15cm}
\section{Deep Inertial Navigation}

\begin{figure*}[t]
\sbox{\bigleftbox}{%
\begin{minipage}[c][5cm][t]{.3\textwidth}
  \vspace*{\fill}
  \centering
  \subfloat[]{
  \includegraphics[width=1.00\textwidth]{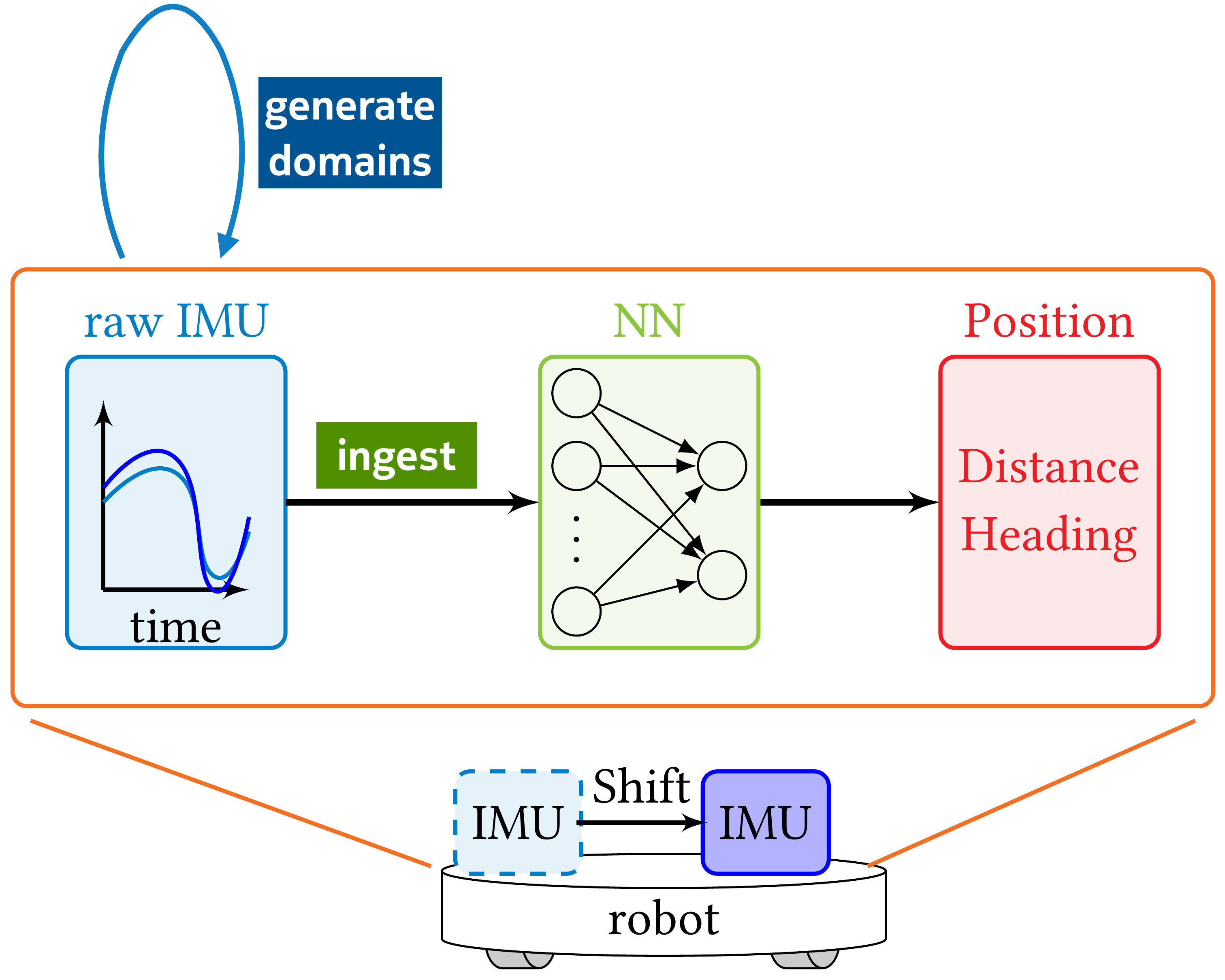}
  \label{fig:diversity_learning}
  } \\
  \vspace{-0.25cm}
  \subfloat[]{
  \includegraphics[width=1.00\textwidth]{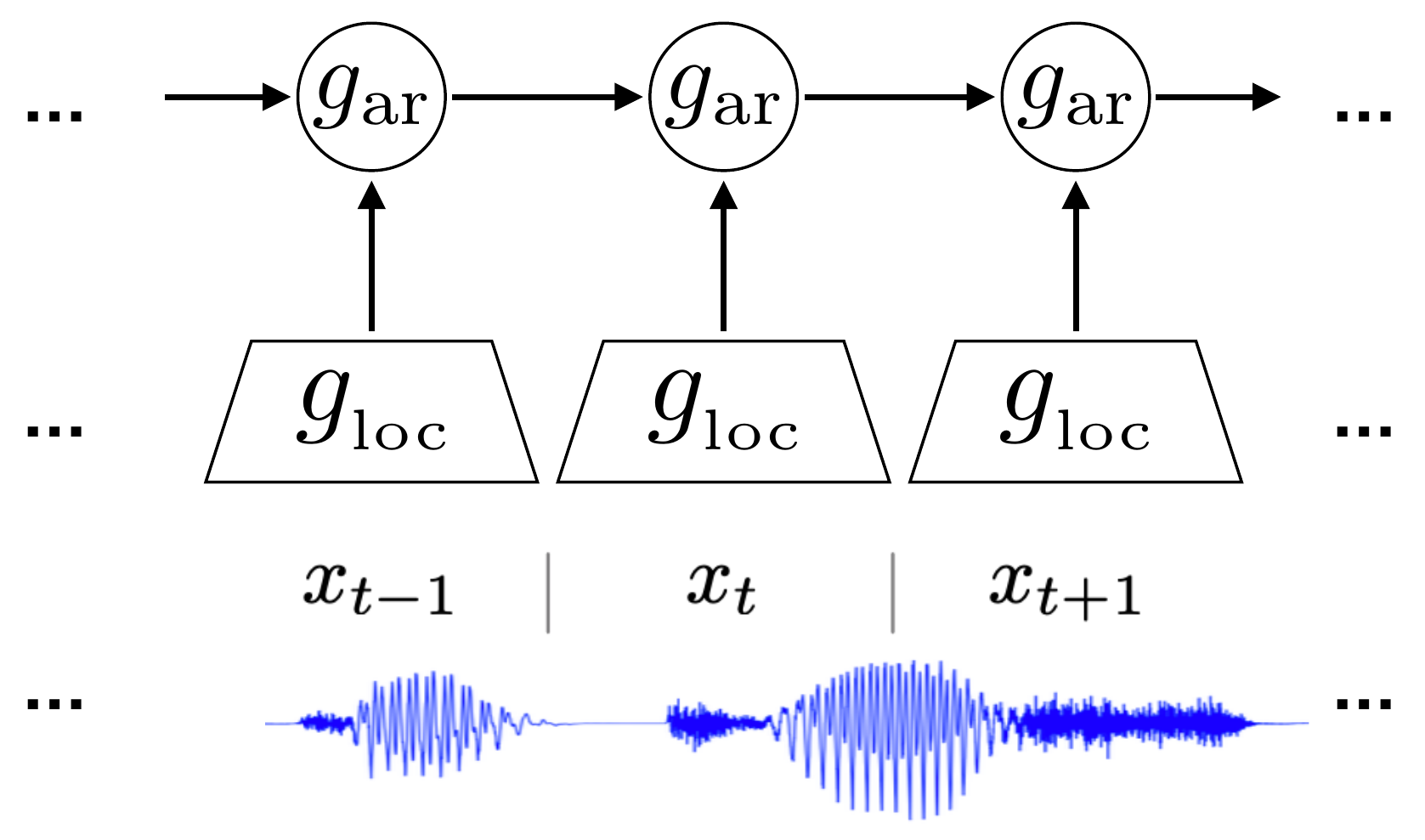}
  \label{fig:encoder_concept}
  }
\end{minipage}
}\usebox{\bigleftbox}%
\begin{minipage}[c][5cm][t]{.7\textwidth}
  \vspace*{\fill}
  \centering
  \vspace{0.95cm}
  \subfloat[]{
  \includegraphics[width=1.00\textwidth, trim=0 0 4cm 0, clip]{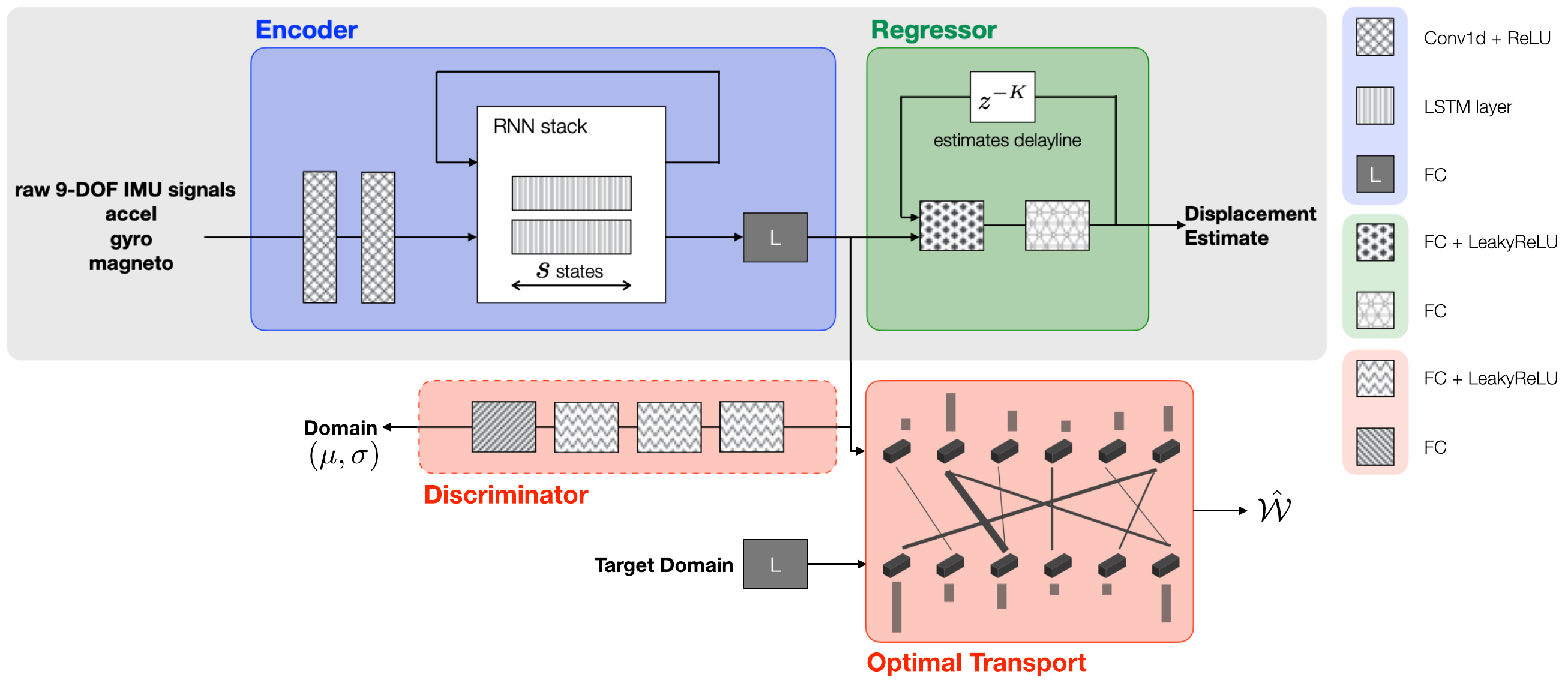}
  \label{fig:nn_architecture}
  }
\end{minipage}%
    \vspace{3.3cm}
    \caption{\small (a) Learning is enriched with data from many sensor positions for better robustness and generalisability.
    (b) Unrolled view of the time-series encoder as proposed in~\cite{Oord19_CPC} comprising a local summaries component $g_{\text{loc}}$ and an autoregressive component $g_{\text{ar}}$.  
    (c) Our proposed deep inertial tracker architecture. The core tracker is composed of an encoder and a regressor. Adapting to a new target domain is handled at training-time by an optimal transport (OT) loss term that measures the dissimilarities of latent spaces, and wherein the thickness of the connections denote proportional mass mapping. In another adversarial incarnation, the core tracker can be alternatively equipped with a discriminator in similar vein to OT in order to enforce domain invariance.}
\end{figure*}

We turn to discuss various design aspects around our end-to-end \ac{DL} inertial tracker.
Our approach is to build a neural architecture that regresses both displacement and heading.
The architecture is further augmented with the \emph{robotic-learning} strategy illustrated in Fig.~\ref{fig:diversity_learning}. The robotic-learning strategy assimilates knowledge from a diversity of sensor positions, thereby enriching the overall learnt model.

\subsection{Representation Learning} \label{sec:nn_architecture}

\noindent \textbf{Architecture.} The core \ac{DL} tracker architecture consists of a pair of encoder $E$ and regressor $R$ as depicted in Fig.~\ref{fig:nn_architecture}. 

The encoder $E$ maps a window of raw 9-DoF inertial signals $\textbf{x}$ into a latent space $E(\textbf{x})$.
In~\cite{Oord19_CPC}, van den Oord et al. propose a time-series encoder that comprises a local summaries component and an autoregressive component, respectively $g_{\text{loc}}$ and $g_{\text{ar}}$ in Fig.~\ref{fig:encoder_concept}. 
The encoder $E$ is inspired by~\cite{Oord19_CPC} and (1) uses a \ac{CNN} to embed a window of \emph{local} raw inertial signals $\textbf{x}$ into a lower dimensional space, (2) feeds the \ac{CNN} embedding into an \emph{autoregressive} \ac{RNN} that tracks the temporal dynamics of inertial navigation, and (3) applies a third learned transformation using a \ac{FC} layer. Thus at a high-level, the architecture of encoder $E$ is a composite \ac{CNN}-\ac{RNN}-\ac{FC} network.

Th encoder $E$ then keeps track of inertial navigation dynamics over $S$ states using a \ac{LSTM} \ac{RNN} variety. Two such \ac{LSTM} layers are stacked for added discriminative power~\cite{Chen19_DnnInertialOdometry}, and whose output is further passed through the \ac{FC} block denoted by the letter $\text{L}$ in Fig.~\ref{fig:nn_architecture}.   

The regressor $R$ maps $E$'s latent space into a displacement estimates vector $\hat{\textbf{y}} = R\left(E(\textbf{x})\right)$. 
The regressor $R$ does so using a cascade of two FC layers, interleaved with a  nonlinearity. 
The regressor $R$ also makes use of previous estimates in order to further enhance its current estimate. This aspect of $R$ constitutes a form of white-box learning. We have found that this explicit feedback is quite effective at enhancing estimation in spite of the implicit feedback already taken place within the LSTM.

\noindent \textbf{Coordinate representation.} The proposed network architecture also supports multiple internal coordinate system representations. That is, two variants for $\hat{\textbf{y}}$ can be realised atop our proposed base architecture
\begin{align}
    (\Delta x, \Delta y) = f_{_\Theta}^{\text{c}}(\textbf{a}, \boldsymbol{\omega}) \\
    (\Delta d, \Delta \phi) = f_{_\Theta}^{\text{r}}(\textbf{a}, \boldsymbol{\omega})
    \label{eq:cartesian_vs_polar_networks}
\end{align}
where $f_{_\Theta}^{\text{c}}$ and $f_{_\Theta}^{\text{r}}$ are the Cartesian and Polar networks, respectively. 
For the Polar network $f_{_\Theta}^{\text{r}}$, the activation \texttt{Tanh} is used for the heading $\Delta \phi$, scaled by $\pi$ for a numerical range $\in [-\pi, \pi]$.

\subsection{Domain Adaptation}

\noindent \textbf{Loss.} The loss function is composed of a) a regression term for displacement estimation and b) an alignment term for generalisability across xSlider domains. 
We follow a joint discriminative learning and latent spaces dissimilarity minimisation procedure~\cite{Bhushan18_DeepJDot} which we detail next.

\begin{figure*}[t]
  \centering
  \subfloat[]{
    \includegraphics{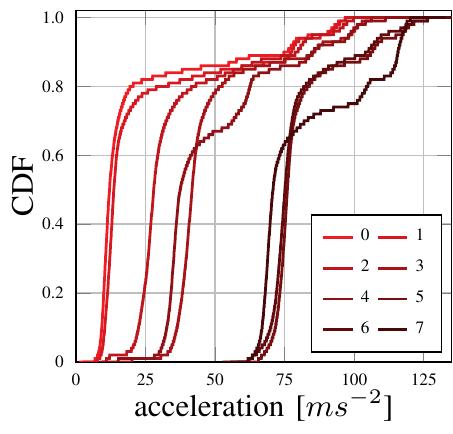}
    \label{fig:raw_accel}
  }
  \subfloat[]{
  \hspace{-0.35cm}
    \includegraphics{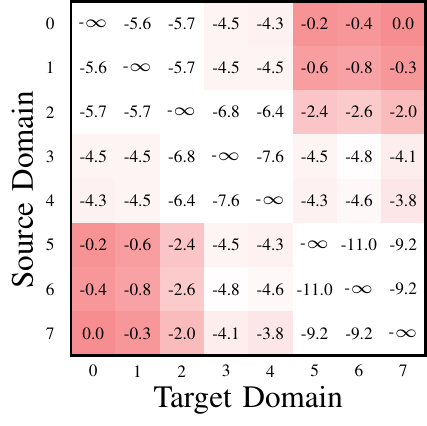}
    \label{fig:raw_wasserstein}
  }
  \subfloat[]{
  \hspace{-0.30cm}
    \includegraphics{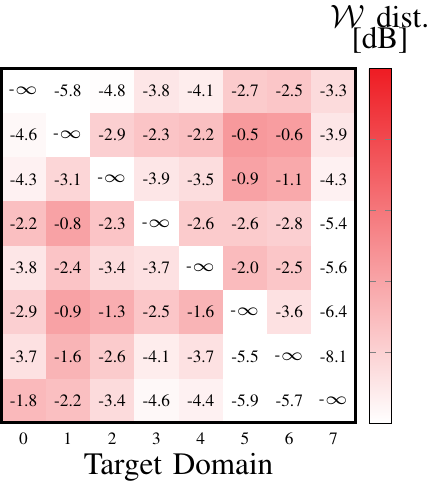}
    \label{fig:latent_wasserstein}
  }
  \subfloat[]{
  \hspace{-0.85cm}
    \includegraphics{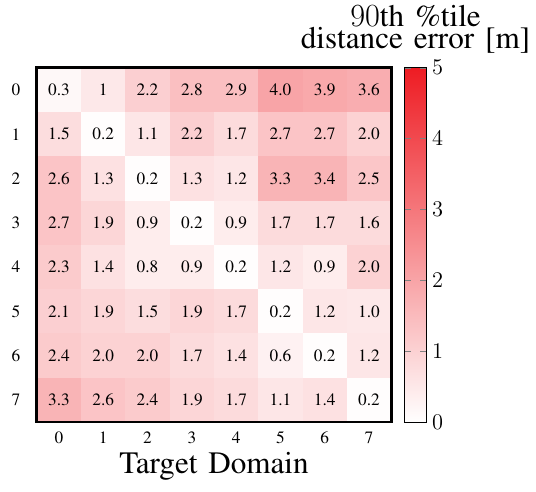}
    \label{fig:inertial_dl_fragility}
  }   
  \vspace*{-0.25cm}
  \caption{Deep learning inertial models are fragile. \protect\subref{fig:inertial_dl_fragility} A mere 1cm change in the sensor position causes a \ac{DL} model trained on an adjacent sensor position to lose significant accuracy. The loss in accuracy is broadly commensurate with the distribution shift seen in raw acceleration data c.f. \protect\subref{fig:raw_accel}~\&~\protect\subref{fig:raw_wasserstein}: \protect\subref{fig:raw_accel} CDFs of the raw 3D acceleration across domains~and~\protect\subref{fig:raw_wasserstein} their corresponding Wasserstein distances. \protect\subref{fig:latent_wasserstein} Wasserstein distances between the latent spaces of per-sensor position models.}
  \label{fig:inertial_shift}
\end{figure*}

For both the Cartesian and Polar coordinate representations, we use the MSE loss
\begin{align}
   \ell_{_\text{MSE}}(\textbf{y}, \hat{\textbf{y}}) =\frac{1}{N} \sum_{n=1}^N \| \textbf{y}_n - \hat{\textbf{y}}_n \|^2
   \label{eq:loss-mse}
 \end{align}
 to compare predictions $\hat{\textbf{y}}_n$ against groundtruth $\textbf{y}_n$.

The alignment term of the loss is computed using the iterative Sinkhorn algorithm as a differentiable proxy to the true Wasserstein distance~\cite{Genevay18_learningGenerativeModelsWithSinkhornDivergences}. The Wasserstein distance quantifies the dissimilarity of two empirical distributions in a geometric sense~\cite{Peyre20_ComputationalOT}. For DA, the objective is to train to minimise the Wasserstein distance between the source and target latent spaces~\cite{Courty17_OtForDomainAdaptation, Bhushan18_DeepJDot}, as highlighted in Fig.~\ref{fig:nn_architecture}. Note that standard Sinkhorn iterations result in biased and slow Wasserstein convergence. Cuturi  et al. prescribe simple but powerful debiasing and regularisation schemes~\cite{Cuturi2013_Sinkhorn,Genevay18_learningGenerativeModelsWithSinkhornDivergences} which we also adopt. Concretely, denote by $\mu$ and $\nu$ the two empirical distributions of the source and target latent space, the regularised Wasserstein distance is given by
\begin{align}
  \hat{\mathcal{W}}(\mu, \nu) = \underset{\text{s.t.} \; \gamma \textbf{1}=\mu, \: \gamma^T \textbf{1}=\nu}{\underset{\gamma \in \Gamma(\mu, \nu)}{\mathrm{argmin}}} \, \langle \, \boldsymbol{\gamma},\textbf{C} \, \rangle_{_\text{F}} - \lambda_{\epsilon} h(\boldsymbol{\gamma})
  \label{eq:wasserstein-regularised} \\
  \textbf{C}_{ij} = \alpha \, \| \mu_i - \nu_j \|^2 + \ell_{_\text{MSE}} \left( \textbf{y}_i^s, R(E(\textbf{x}_j^t)) \right) \notag
\end{align}
where $\boldsymbol{\gamma}$ is the probabilistic coupling between $\mu$ and $\nu$, $\Gamma$ is the space of the joint probability distributions with marginals $\mu$ and $\nu$, $\textbf{C} \ge 0$ is a cost matrix $\in \mathbb{R}^{n_\mu \times n_\nu}$ representing the pairwise distances between source and target latent distributions, $\langle \cdot, \cdot\rangle_\text{F}$ is the Frobenius dot product, $h$ is the Shannon entropy, and $\lambda_{\epsilon}$ is a regularisation coefficient. We use an $\text{L}2$ moment for $\textbf{C}$. $\textbf{C}$ further \emph{mixes} in pairwise distances of the source and target label\footnote{Note that a so-called surrogate label~\cite{Bhushan18_DeepJDot} is used for a target domain with unavailable groundtruth.} distributions, scaled by hyperparameter $\alpha$. The superscripts $(\cdot)^s$ and $(\cdot)^t$ denote the source and target domains, respectively. 

For a batch of data $(\textbf{x}_i^s, \textbf{y}_i^s)_{i=1}^I$ and $(\textbf{x}_j)_{j=1}^J$, the loss can then be written as a sum $\mathcal{L} = \mathcal{L}_\text{regression} + \mathcal{L}_\text{alignment}$, where
\begin{align}
  \mathcal{L}_\text{alignment} &= \hat{\mathcal{W}}(\mu, \nu) - 0.5 \left( \hat{\mathcal{W}}(\mu, \mu) + \hat{\mathcal{W}}(\nu, \nu) \right) \label{eq:loss-alignment},\\[-3pt] 
  \vspace{-0.2cm}
  \mathcal{L}_\text{regression} &= \frac{1}{I}\sum_{i=1}^I\ell_{_\text{MSE}} \left( \textbf{y}_i^s, R(E(\textbf{x}_i^s)) \right)\notag \\[-3pt]
  &+ \sum_{j=1}^J\boldsymbol{\gamma}_{i,j} \, \ell_{_\text{MSE}} \left( \textbf{y}_i^s, R(E(\textbf{x}_j^t)) \right). \label{eq:loss-regression}
\end{align}
The alignment term in Eq.~\eqref{eq:loss-alignment} debiases Wasserstein estimates, while the regression term in Eq.~\eqref{eq:loss-regression} mixes the source MSE loss with a \emph{transported}\footnote{The transport coupling $\boldsymbol{\gamma}$ is a by-product of computing $\mathcal{W}$.} version of the target MSE loss due to the unavailability of labels in the target domain.

\noindent \textbf{Training.} With the architecture and loss described, the final training employs the DeepJDOT alternating procedure~\cite{Bhushan18_DeepJDot} that jointly optimises for displacement regression and domain alignment.

\SetKwProg{Fn}{}{}{}
\SetKwFunction{FRecurs}{FnRecursive}%
\SetAlgoNoLine
\SetKwInOut{Input}{input}
\SetKwInOut{Output}{output}

\vspace{-0.25cm}
\begin{algorithm}

  \Input{labelled source domain data $(\textbf{X}^s, \textbf{Y}^s)$ \\ unlabelled target domain data $\textbf{X}^t$}
  
  \For{each batch of source and target in training set}{
    with fixed $E$ \& $R$, solve for $\boldsymbol{\gamma}$;     \\
    with fixed $\boldsymbol{\gamma}$, update $E$ \& $R$;        \\
    backpropagate combined loss;
  }

\caption{Joint regression and alignment training\label{algo:training_pseudocode}}
\end{algorithm}

\subsection{Insight -- Wasserstein Distance Analysis}

\begin{figure*}
  \subfloat[]{
    \includegraphics{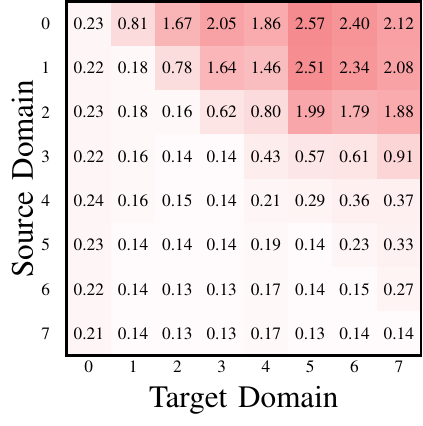}
    \label{fig:OT_DistError}
  }
  \hspace{-0.33cm}
  \subfloat[]{
    \includegraphics{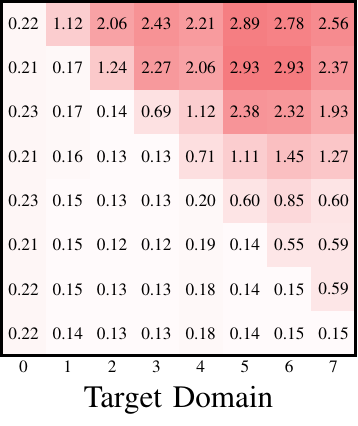}
    \label{fig:Aug_DistError}
  }
  \hspace{-0.33cm}
  \subfloat[]{
    \includegraphics{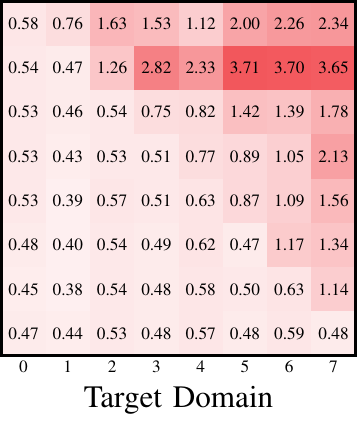}
    \label{fig:Pcida_DistError}
  }
  \hspace{-0.33cm}
  \subfloat[]{
    \includegraphics{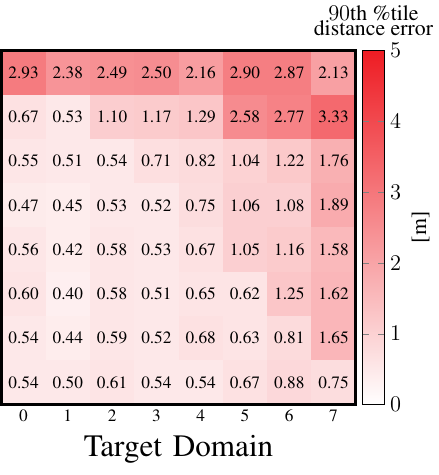}
    \label{fig:Cida_DistError}
  }
  \vspace*{-0.2cm}
  \caption{$90\text{th}$\%tile distance error for 4 experiments. \protect\subref{fig:OT_DistError} OT DA. \protect\subref{fig:Aug_DistError} multi-domain augmentation. \protect\subref{fig:Pcida_DistError} PCIDA. \protect\subref{fig:Cida_DistError} CIDA.}
  \label{fig:DistErrorMatrix}
\end{figure*}

Here we shed light on our research question 1 from Section~\ref{Sec:Introduction}, namely: \emph{What is the effect of different sensor positions on inertial signals?}

Fig.~\ref{fig:raw_accel} shows the $\text{L}2$-norm of the raw 3D acceleration for our xSlider domains. We can see significant shifts in the acceleration CDFs as a function of the sensor position during navigation. 
These shifts can be quantified using the 1D Wasserstein distance\footnote{$\mathcal{W}(\mu, \nu) = \int_0^1 c \big( \, | F_{\mu}^{-1}(x) - F_{\nu}^{-1}(x) | \, \big) dx$, where $\mu$ and $\nu$ are empirical distributions, $F$ is the CDF function, and $c$ denotes a cost function.} as shown in Fig.~\ref{fig:raw_wasserstein}.

We train one DL model\footnote{whose architecture is treated in Section~\ref{sec:nn_architecture}} per xSlider domain using the \ac{IMU} measurements and groundtruth labels, and test on \ac{IMU} data from all 8 domains.
The $90\text{th}$ \%tile performance of these per-domain models is analysed in Fig.~\ref{fig:inertial_dl_fragility}. 
It can be readily seen that considerable performance degradation is incurred even by adjacent domains as little as $1$cm apart from the source domain on which the model has been trained.
Further, inspecting Figs.~\ref{fig:raw_wasserstein}~\&~\ref{fig:inertial_dl_fragility}, the degradation in performance is broadly commensurate with the raw acceleration shifts.
\emph{To what can we attribute such fragility in these per-domain models?}

Fig.~\ref{fig:latent_wasserstein} measures the shifts between the latent spaces\footnote{Box marked with letter $\text{L}$ in Fig.~\ref{fig:nn_architecture}} of these per-domain models using the Wasserstein distance~\cite{Peyre20_ComputationalOT}.
We note that Fig.~\ref{fig:latent_wasserstein} follows too the shift pattern in Figs.~\ref{fig:raw_wasserstein}~\&~\ref{fig:inertial_dl_fragility}.
That is, the latent shifts follow broadly the shifts in the raw xSlider dataset.
Our model, therefore, \emph{preserves the structure of the underlying data}.

This empirical analysis of Wasserstein distances suggests that inducing incremental shifts in sensor position corresponds to: a) quantifiable shifts in raw inertial signal distributions, and b) quantifiable shifts in inertial distributions as seen through the lens of the latent space.
This observation motivates us to attempt to devise an inertial learning scheme that can benefit from the structure of the latent space.

\vspace{-0.15cm}
\section{Experiments} \label{sec:experiments}

We evaluate our neural architectures and algorithms on xSlider: the first domain-\emph{indexed} inertial navigation dataset. 
We begin by discussing our evaluation setup and baselines.

\subsection{Setup}
We segment our xSlider dataset into navigation sessions, each 20-second in duration.
We obtain about 3.8k sequences for training, and 950 for testing. 
The input corresponds to 9-DoF, each 3D.
The the groundtruth labels are 2D coordinates provided by robot's SLAM system, with lidar contributing the lion's share of label accuracy.

We conduct a thorough hyper-parameter search along with architectural exploration using the \ac{NNI} tool by Microsoft~\cite{NNI}. 
For our experiments we train on eight GeForce RTX 2080 Ti GPUs.
For all experiments, we report the average performance of 8 stochastically trained models.  
A further detailed description is available in supplementary material.

\subsection{Baselines}
We begin by discussing our implementations of a number of baselines: (i) \& (ii) for general inertial navigation, and (iii) \& (iv) for continuous domain adaption.

\noindent \textbf{(i) Conventional fusion.}
We use Bosch's BNO055 on-device fusion to obtain heading and linear acceleration (i.e. with gravity subtracted) estimates~\cite{IMUManual}. We then perform double integration to estimate displacement.

\noindent \textbf{(ii) RoNIN.}
RoNIN is a state-of-the-art neural inertial navigation system~\cite{Herath20_Ronin}. RoNIN open-source implementation supports three backbone architectures: Residual Neural Network (ResNet), LSTM, or Temporal Convolutional Network (TCN)~\cite{RoNIN_implementation}. 
We modify RoNIN's LSTM variant for xSlider, as it performed best. 
RoNIN relies on auxiliary orientation estimates~\cite{Herath20_Ronin}. 
We, therefore, use the highly accurate heading estimation from the Bosch sensor for our RoNIN xSlider implementation. 

\noindent \textbf{(iii) Vanilla augmentation.}
Data augmentation is one straightforward and widely used approach to handling source domain shift. For instance, synthetic randomly rotated images are routinely used in computer vision (CV) training pipelines. We include a multi-domain augmentation baseline on xSlider to understand DA’s advantage over simplistic augmentation.

\noindent \textbf{(iv) Continuously-indexed domain adaptation.}
CIDA is the first work to advocate for discriminative learning and adaptation across a continuous set of domains~\cite{Wang20_CIDA}. We adapt CIDA and its probabilistic variant PCIDA for our xSlider (cf. Fig.~\ref{fig:nn_architecture}) to benchmark OT against this state-of-the-art DA method.

\vspace{-0.25cm}
\section{Results} \label{sec:results}
We present next results obtained using 4 baselines + OT DA, all as evaluated on our precision robotics dataset: xSlider.

\subsection{Navigation}

\begin{figure}[h]
    \centering
    \includegraphics{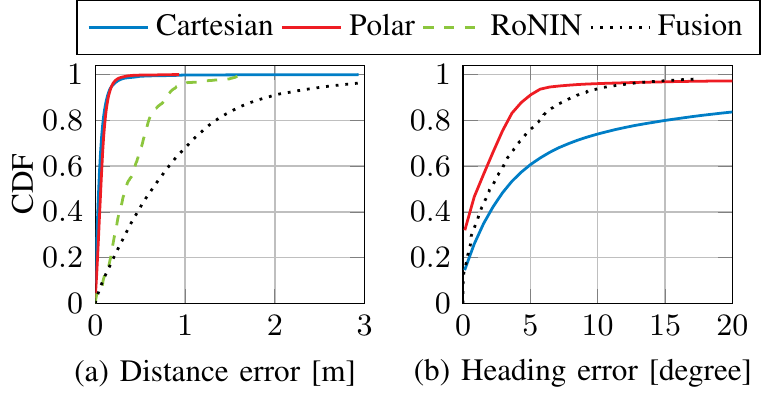}
    \vspace*{-0.25cm}
    \caption{Navigation performance.}
    \label{fig:navigation_vs_baselines}
\end{figure}

Fig.~\ref{fig:navigation_vs_baselines} depicts the navigation performance of our DL tracker and compares it to RoNIN and fusion.
On the xSlider, the 90$\text{th}$ $\%$ile distance errors are 80cm and 2m, respectively for RoNIN and fusion.
As discussed, RoNIN relies on the quite accurate heading estimates from fusion, which are under \SI{9}{\degree} at the 90$\text{th}$ $\%$ile.
Our DL tracker performs much better with around 10cm distance error at the 90$\text{th}$ $\%$ile irrespective of the label representation. 
However, the Polar variant has much better heading estimate performance compared to Cartesian, which is on par with fusion i.e. \SI{9}{\degree} at the 90$\text{th}$ $\%$ile. 
The performance of our DL tracker suggests that the combination of local convolutions and long autoregressive tracking represents an effective base architecture for time-series inertial navigation. 

Due to the large number of models trained for DA experiments, we will use the Cartesian variant for benchmarking because of computational advantage during training over Polar.

\subsection{Adaption} \label{sec:adapation}
We investigate here how much labelled data various DA techniques require before beginning to generalise to unlabelled data. 
The analysis uses the $90\text{th}$\%ile distance error as a single metric proxy to tracking performance. 
The experiments are devised as follows. 
Starting with domain index 0, the number of consecutive domains designated as source is incrementally increased from 1 to 8. 
For example, using 3 domains as source data entails a) supervised learning on xSlider indices 0, 1, and 2, and b) unsupervised adaptation on xSlider indices 3 to 7. 
We repeat this experiment for our 3 contender DA techniques: CIDA, PCIDA, and OT. 
We further include a multi-domain augmentation baseline. 

Fig.~\ref{fig:DistErrorMatrix} depicts the distance error matrices for the 4 methods.
We begin by inspecting Fig.~\ref{fig:Aug_DistError} for multi-domain augmentation, whose upper triangular section corresponds to unseen target domains.
\emph{It is evident that multi-domain augmentation has the worse performance on unseen target domains owing to lack of training-time domain invariance optimisation.} 
In contrast, the upper triangular section of OT in Fig.~\ref{fig:OT_DistError} enjoys the best performance out of the other 4 methods.
Specifically, \emph{beginning at index 3, the OT-adapted inertial tracker is able to transfer its learnt knowledge to other robotically-indexed domains completely without labels (i.e. no groundtruth).}  
Such phenomenon is best understood referring back to Fig.~\ref{fig:TopPictureRoborock} and domain annotations therein. 
Concretely, the first 4 xSlider indices (0 to 3) gradually approach its centre; meaning, once knowledge from these 4 indices is assimilated, \emph{OT is able to extend this knowledge to remaining domain indices by virtue of geometrical symmetry}.
CIDA and PCIDA, on the other hand, suffer from longer error tail despite some modest improvements over multi-domain augmentation in their upper triangles.
It is also timely to point out that \emph{OT is at least $10\times$ faster to train than adversarial CIDA and PCIDA}.
It would seem that \emph{OT's geometric nature is a good match to dealing with sensor position-induced inertial domain shifts}.
That said, perhaps we can ``tweak'' CIDA further (e.g. its discriminator's $\text{L}2$ loss) to begin to approach OT's performance.

\subsection{Generalisation} \label{sec:generalisation}

 \begin{figure}[t]
     \centering
     \includegraphics{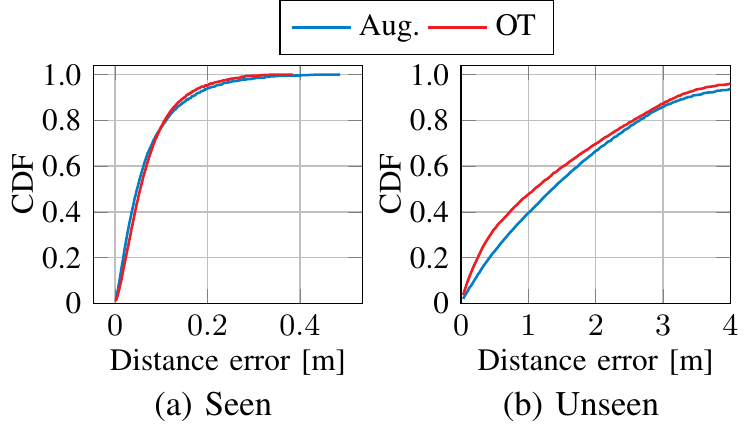}
     \vspace{-0.2cm}
     \caption{OT outperforms augmentation in distance error performance both on seen and unseen domains} 
     \label{fig:generalisation_cdfs}
 \end{figure}
OT and augmentation have emerged as the best knowledge assimilators from Sec.~\ref{sec:adapation}. 
We thus turn to examine their performances more closely on seen and unseen domains. 
To this end, Fig.~\ref{fig:generalisation_cdfs} evaluates the distance error of OT and augmentation on: a) seen xSlider domains 0 to 7, and b) unseen data from the stationary IMU1 cf. Fig.~\ref{fig:TopPictureRoborock}.
In Fig.~\ref{fig:generalisation_cdfs}a, we can see marginal OT improvements in the error long tail on seen domains. 
We similarly see in Fig.~\ref{fig:generalisation_cdfs}b consistent OT enhancements to performance on a totally unseen domain compared to augmentation.
Both observations suggest that OT DA does aid in coherently assimilating the knowledge contained in the indexed xSlider dataset.

\vspace{-0.25cm}
\section{Related work}
Closest to our work, Oxford University researchers propose \ac{RNN}-based inertial tracking for regressing both displacement and heading, which outperforms standard dead reckoning~\cite{Chen18_OxIOD}.
However, they employ a bidirectional LSTM which incurs latency i.e. \emph{acausal} estimation. 
We, in contrast, rely on strictly causal estimation. 
More device-friendly inertial tracking DL architectures were proposed in~\cite{IMULet, Chen20_DeepOndeviceInference}.
A number of recent DL inertial tracking results have been reported, either end-to-end~\cite{Herath20_Ronin,Chen19_DnnInertialOdometry,EndToEnd}, or by aiding state-space formulations~\cite{Liu20_Tlio, Brossard20_AI-IMU}.
In either formulation, prior works acknowledge the difficulty of generalising DL-based trackers beyond the scope of training data~\cite{Liu20_Tlio}, especially for the more challenging wheeled motion profiles~\cite{Chen19_DnnInertialOdometry}.

\vspace{-0.20cm}
\section{Conclusion}
In this paper, we study the effect of sensor position during wheeled robotic navigation on inertial signals and their learnt latent spaces.
We propose new neural architectures and algorithms to assimilate knowledge from an \emph{indexed} set of sensor positions.
We provide an early quantitative evidence for the benefits of inertial learning using a diversity of sensor positions.

Currently, state-of-the-art visual inertial odometry systems require careful calibration of the visual and inertial sensors, including precise knowledge of their placement on the robot~\cite{ORB-SLAM3}.
This work takes a first step towards alleviating such a hard requirement through either: a) unsupervised adaptation towards an arbitrarily placed IMU (Sec.~\ref{sec:adapation}), or more ambitiously b) generalisation out-of-the-box to unseen IMU signals (Sec.~\ref{sec:generalisation}).
The ultimate aim is to reduce the overhead cost by removing the need for calibration altogether.
That is, our end-to-end learning has the potential to reduce error growth without explicitly requiring the estimation of IMU biases (e.g., in contrast to~\cite{ORB-SLAM3}).
This ambition is likely to require yet more elaborate robotic-learning schemes.

\bibliographystyle{IEEEtran}
\bibliography{IEEEabrv,biblio}

\newpage
\section*{APPENDIX}
\section{Design \& Fabrication}

In this section we document the design and fabrication of the xSlider robotic system.

\subsection{xSlider system}\label{Sec:MeasSys}

The xSlider is a flexible and versatile measurement platform for automatically creating a large amount of labeled data. It is based on a vacuum cleaner robot consisting of: a two-dimensional \ac{LiDaR}, ultrasonic distance sensors, wheel-tick sensors, multiple \ac{IR} sensors, and an \ac{IMU}. 
These modalities are fused together---via a Kalman filter-based \ac{SLAM} algorithm---to generate pseudo-groundtruth for our subsequent experiments.
The \ac{SLAM} accuracy is better than \SI{1}{\centi\metre}, as per the evaluation conducted by Hoffmann et al.~\cite{Hoffmann2020}.
Such accuracy satisfies the requirements of most localization applications, including the geometry-aware IMU DL we study in this paper.
Fig.~\ref{fig:measurement_system} depicts the xSlider measurement system we utilise throughout our measurement campaign. 
It consists of: (1) an robot for the pseudo-groundtruth, (2) a platform  mounted on top of the robot to host additional sensors, (3) a \ac{MQTT} broker with a database for data acquisition, and (4) an \ac{NTP} server for time synchronization between various sensors.
In terms of hardware, the platform was fabricated in-house from 3D printed parts and a LEGO building plate. 
On the software side, a \ac{MQTT} broker and a \ac{NTP} server form the core of the data acquisition system. 
Specifically, each sensor synchronizes its time-clock with the local \ac{NTP} server, which allows for an accuracy below \SI{1}{\milli\second}~\cite{Milis1991}. 
The synchronized measurements are then streamed to the \ac{MQTT} broker. 
By virtue of \ac{MQTT} and the back-end database, our sensor system is modular and extensible. 
We leverage a Roborock S50 from Xiaomi as base robot platform.
The robotic platform streams its pseudo-groundtruth positions to the \ac{MQTT} broker with a \SI{5}{\hertz} update rate. 
A powerbank of \SI{10}{\ampere\hour} is used to power the sensors.

For our empirical \ac{DA} measurements, we have further built a linear slider subsystem, based on printed 3D parts and stepper motors. 
Such mechanically-indexed slider has been shown to achieve a high movement accuracy~\cite{Arnold2017}. 
As shown in Fig.~\ref{fig:TopPictureRoborock}, the slider facilitates the translation of IMU2 position programmatically along a designated axis.
The stationary IMU1 acts as a reference \emph{paired}\footnote{this is a \ac{DA} technical term} measurement to that of IMU2.

We utilize \ac{IMU} modules from Adafruit~\cite{AdaFruit} equipped with Bosch BNO055 inertial chipsets~\cite{IMUManual}. 
Each \ac{IMU} module is connected to an ESP32 \SI{2.4}{\giga\hertz} WiFi chipset over an I2C interface.
We have found that this setup achieves an average sampling rate of \SI{70}{\hertz} for all two\footnote{from the paired IMUs} 9-\acp{DoF} ($x/y/z$-acceleration, $x/y/z$- magnetometer, $x/y/z$-rotation rate). 
The slider is also connected with an ESP32 which can be controlled at run-time over \ac{MQTT} commands, allowing it to move with an accuracy of \SI{1}{\milli\metre}.  
Note the slider positional annotations are depicted in \ref{fig:TopPictureRoborock}. 
The center of mass of the robot lies between the $3$rd and $4$th domain indices. 
The domain index is fixed throughout each measurement session in order to facilitate analysis and comparison.
Measurement sessions also cover the same area for added consistency.
We have made added effort to control for these factors during our empirical data collection campaign in order to foster much needed insights and understanding around sensor position-induced IMU variabilities.

\begin{figure}[t]
    \centering
    \includegraphics{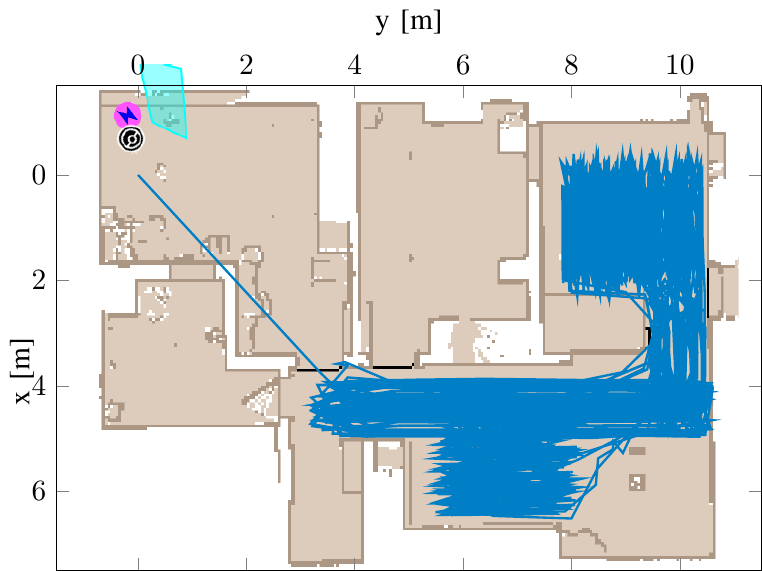}
     \caption{Flat floorplan in which data collection campaign took place.} 
    \label{fig:MeasPath}
    \vspace{-0.4cm}
\end{figure}

\subsection{Pseudo groundtruth}\label{Sec:MeasSysGT}

Fig.~\ref{fig:MeasPath} depicts the measurement area within a flat's floorplan located in Stuttgart, Germany. 
The robot trajectories are highlighted in blue. 
The floorplan has been automatically created by the robotic \ac{SLAM} algorithm. 
The measurement campaign was conducted over 8 days.
Specifically, IMU2 was mechanically-indexed along $8$ slider positions on the $x$-axis as shown in Fig.~\ref{fig:TopPictureRoborock}.
These $8$ slider positions correspond to offsets between \SI{-0.4}{\centi\metre} to \SI{6.6}{\centi\metre}, as referenced to the robot's center of mass.
The combined slider positions amount to exactly \SI{7}{\centi\metre} translation of mechanically-indexed IMU domains.
Note that we have also calibrated for the precise position of the center of the \ac{IMU} breakout board.

Coordinate translation of the \ac{SLAM} pseudo groundtruth is needed in order to solve for \acp{IMU} positions.
Because of the peculiarities of the robot coordinate system, the robot may spin in-place and report a stationary position while causing the mounted \acp{IMU} to experience accelerations and rotational speeds. 
In order to correct for these effects, we devised the following. 
Firstly, the coordinate system of the \ac{SLAM} calculates the position based on an anchor point located at the front of the robot.
The anchor point is translated to the center of mass of the robot. 
The center of mass is the origin of the $x$ and $y$ axes in Fig.~\ref{fig:TopPictureRoborock}.\footnote{Due to vacuuming chamber and asymmetries in the construction of the robot, the center of mass does not coincide with the geometric centre. Thus, our choice of center of mass references \ac{DA} experiments to acceleration and not rotation.}
The \ac{SLAM} groundtruth coordinates $(x_{_\text{SLAM}}, y_{_\text{SLAM}})$ and orientation $\phi_{_\text{SLAM}}$ are referenced to the robot coordinate system. Denote by $(\hat{x}_{_\text{SLAM}}, \hat{y}_{_\text{SLAM}})$ and  $\hat{\phi}_{_\text{SLAM}}$ a translated reference coordinate and orientation system, respectively. 
The offset angle $\phi_{_{\text{IMU},\ell}}$ and coordinates $(x_{_{\text{IMU},\ell}}, y_{_{\text{IMU},\ell}})$ of the \acp{IMU} ($\ell \in [1,2]$) were measured.
Finally, these measured values are used to translate the position of the \acp{IMU} to the global coordination system, according to

\begin{align}
\label{eqn:IMU Positions}
\begin{split}
r_{_{\text{IMU},\ell}} &=  \sqrt{x^2_{_{\text{IMU}, \ell}} +  y^2_{_{\text{IMU},\ell}}}
\\
 \hat{x}_{_{\text{IMU},\ell}} &= \hat{x}_{_\text{SLAM}} + \text{Re}\Bigl\{ r_{_{\text{IMU},\ell}} \; e^{j\left(\hat{\phi}_{_\text{SLAM}}+\phi_{_{\text{IMU},\ell}}\right)} \Bigr\}
\\
 \hat{y}_{_{\text{IMU},\ell}} &= \hat{y}_{_\text{SLAM}} + \text{Im}\Bigl\{ r_{_{\text{IMU},\ell}} \; e^{j\left(\hat{\phi}_{_\text{SLAM}}+\phi_{_{\text{IMU},\ell}}\right)} \Bigr\}
\end{split}
\end{align}

\subsection{Construction.} 
For systematic \ac{DA} experiments grounded in physics, we use the xSlider system to precisely translate the position of the IMU while minimizing mechanical vibrations. 
To this end, the mechanical assembly is composed of a mounting rail with a ball-bearing slider driven by a timing belt. The timing belt is attached to a stepper engine, allowing micro-step controls as to realize a movement accuracy within \SI{}{\milli\metre}s. 
The mounting rail is screwed to a 3D printed mounting frame, which is in turn glued onto LEGO terminal blocks. 
This allows for easy removal and reconfiguration of parts. 
3D printing is done via the Fused Deposition Modeling (FDM) method. 
All CAD files necessary for 3D printing the mounting structure and its part will be made available upon dataset release.


\end{document}